\pdfoutput=1 

\documentclass{article}

\usepackage{arxiv}
\usepackage{physics}
\usepackage{qcircuit}
\usepackage[utf8]{inputenc} 
\usepackage[T1]{fontenc}    
\usepackage{hyperref}       
\usepackage{url}            
\usepackage{booktabs}       
\usepackage{amsfonts}       
\usepackage{nicefrac}       
\usepackage{microtype}      
\usepackage{bbold}
\usepackage{subcaption}
\usepackage{fullpage}
\usepackage{times}
\usepackage{enumitem}
\usepackage{fancyhdr,graphicx,amsmath,amssymb}
\usepackage{mathtools} 
\usepackage[ruled,vlined]{algorithm2e}
\usepackage{nicefrac}
\usepackage{floatrow}
\usepackage{xparse}
\usepackage{xcolor}

\newsavebox{\twosubbox}
\usepackage[titletoc,title]{appendix}

\usepackage{tabularx} 
\usepackage{multirow}

\newcommand{\appendixhead}%
{\textbf{\huge Appendices}
\vspace{0.25in}}

\newfloatcommand{capbtabbox}{table}[][\FBwidth]

\definecolor{maiblue}{rgb}{0, 0., 0.69}
\hypersetup{colorlinks=true, citecolor=maiblue, pdffitwindow=true, linkcolor=maiblue, urlcolor=maiblue}

\definecolor{Gray}{gray}{0.925}

\usepackage{graphicx}
\usepackage{tabularx} 
\usepackage{multirow}
\usepackage{booktabs}
\usepackage[mode=math]{siunitx}
\usepackage{colortbl}
\usepackage{booktabs}
\usepackage{hyperref}

\definecolor{upforestgreen}{rgb}{0.0, 0.27, 0.13}

\title{Quantum Ensemble for Classification}

\author{
  Antonio Macaluso\\
  German Research Center for Artificial intelligence (DFKI)\\
  Agents and Simulated Reality Department \\
   \texttt{antonio.macaluso@dfki.de} \\
   \And
    Luca Clissa \\
    Department of Physics and Astronomy\\
    University of Bologna, Italy\\
    Istituto Nazionale di Fisica Nucleare (INFN), Italy \\
    \texttt{luca.clissa2@unibo.it} \\
  \AND
      Stefano Lodi \\
    Department of Computer Science and Engineering \\
    University of Bologna, Italy\\
    \texttt{stefano.lodi@unibo.it} \\
    \And
    Claudio Sartori \\
    Department of Computer Science and Engineering \\
    University of Bologna, Italy\\
    \texttt{claudio.sartori@unibo.it} \\
}

\begin{document}

\maketitle

\begin{abstract}

A powerful way to improve performance in machine learning is to construct an ensemble that combines the predictions of multiple models. Ensemble methods  are often much more accurate and lower variance than the individual classifiers that 
make them up, but have high requirements in terms of memory and computational time. In fact, a large number of alternative algorithms is usually adopted, each requiring to query all available data. 

We propose a new quantum algorithm that exploits quantum superposition, entanglement and interference to build an ensemble of classification models. Thanks to the generation of the several quantum trajectories in superposition, we obtain $B$ transformations of the quantum state which encodes the training set in only $log\left(B\right)$ operations. This implies an exponential growth of the ensemble size while increasing linearly the depth of the correspondent circuit. Furthermore, when considering the overall cost of the algorithm, we show that the training of a single weak classifier impacts additively to the overall time complexity rather than multiplicatively, as it usually happens in classical ensemble methods.

We also present small-scale experiments on real-world datasets, defining a quantum version of the cosine classifier and using the IBM qiskit environment to show how the algorithms work.

\end{abstract}


\section{Introduction}

Quantum Computing (QC) can achieve performance orders of magnitude faster than the classical counterparts, with the possibility of tremendous speed-up of complex computational tasks \cite{grover1996fast, shor1999polynomial, PhysRevLett.103.150502}. 
Thanks to the quantum mechanical principles of superposition and entanglement, quantum computers can achieve vast amounts of parallelism without needing the multiple replicas of hardware required in a classical computer. 
One of the most relevant fields in which QC promises to make an impact in the near future is machine learning (ML). 
Quantum Machine Learning (QML)  is a sub-discipline of quantum information processing devoted to developing quantum algorithms that learn from data in order to improve existing methods. 
However, being an entirely new field, QML comes with many open challenges \cite{aaronson2015read}.

One of the most studied problems in QML is classification, where an algorithm is trained on 
data 
whose category of the target variable is known. According to Dietterich's definition \cite{dietterich2000ensemble}, a classifier is a hypothesis about the true function $f$ which allows estimating a target variable $y$, given a vector of features $x$. 
Among the multiple alternatives to build a classifier, a well-known approach is ensemble methods, where a large number of hypotheses is combined 
by averaging or voting rules to classify new examples.
Despite the absence of a unified theory, there are many theoretical reasons for combining multiple learners, e.g. reducing the prediction error by decreasing the uncertainty on the estimates, as well as empirical evidence of the effectiveness of this approach \cite{doi:10.1080/095400996116839, domingos2000bayesian}.

\section{Background}

When trying to predict a target variable using any ML model, the main causes of the difference in actual and predicted values (Expected Prediction Error or \textit{EPE}) are noise, bias and variance \cite{hastie2005elements}. 
The \textit{noise} component, also known as \textit{irreducible error}, is the variance of the target variable around its true mean. This error is due to the intrinsic uncertainty of the data, so it cannot be avoided no matter how well the model works.
The \textit{bias}, instead, is linked to the particular learning technique adopted, and it measures how well the method suits the problem. 
Finally, the \textit{variance} component measures the variability  of the learning method around its expected value.
In light of this, in order to improve the performance of any ML technique, one has to try to reduce one or more of these components.

\subsection{Ensemble Learning}\label{sec:ensemble_learning}

The idea of ensemble learning is to build a prediction model by combining the strengths of a collection of simpler base models to reduce the \textit{EPE}.
A necessary and sufficient condition for an ensemble to outperform any of its members is that the single models are \textbf{accurate}, in the sense that they have an error rate better than random guessing, and \textbf{diverse}, which means that the individual models make different errors given the same data points \cite{58871}. 

There exist several ways to build ensemble methods, each designed to tackle a specific component of the \textit{EPE}.
In \textit{Boosting},  the idea is to exploit a committee of weak learners that evolves over time. In practice, at each iteration a new weak learner is trained with respect to the error of the whole ensemble. This mechanism allows getting closer and closer to the true population values, thus reducing the bias.
\textit{Randomisation} methods consist in estimating the single base model with a randomly perturbed training algorithm. This alteration worsens the accuracy of the individual learners,
but reduces the ensemble variance thanks to the combination of a large number of randomised models.
Unlike the other methods, this approach is applicable also to stable learners, thus enlarging the plethora of methods it applies to.
Another approach is \textit{Bagging}. In this case, the same model is fitted to different training sets, 
thus creating a committee of independent weak learners.
The individual votes are then averaged to obtain the ensemble prediction.  
This approach decreases the \textit{EPE} by reducing the variance component 
so the more classifiers are included (i.e., the larger the size of the ensemble), the more significant is the reduction.

In practice, bagging reduces to computing several  predictions $\hat{f}_{1}(x), \hat{f}_{2}(x), \dots , \hat{f}_{B}(x)$ using $B$ different training sets, which are then averaged to obtain a single model with lower variance:
\begin{align}
    \hat{f}_{\text{bag}}(x) = \frac{1}{B} \sum_{b=1}^T \hat{f}_{b}(x).
\end{align}
Although this approach guarantees a lower uncertainty in prediction, it is not practical in its theoretical formulation,
due to the lack of multiple training sets. To overcome this issue, bootstrap procedure \cite{efron1994introduction} can be employed, that takes repeated samples from the available data and generates  $B$ different bootstrapped training sets. The learning algorithm is then trained on the \textit{b-th} bootstrapped observations to get $B$ different predictions $\hat{f}_{b}(x)$.
The difference between the bootstrap and the idealised procedure is the way the training sets are derived. Instead of obtaining independent datasets from the domain, the initial training set is perturbed as many times as the number of weak classifiers to aggregate. The generated datasets are certainly not independent because they are all based on the same training set.
Nonetheless, empirical findings  suggest that bagging is still able to produce combined models that often significantly outperform individual learners, and that anyway are never substantially worse \cite{doi:10.1080/095400996116839}.

\subsection{Related works}

Recently, the idea of a quantum ensemble has been investigated \cite{schuld2018quantum, abbas2020quantum}. 
In this case, the construction of the ensemble corresponds to three different stages: \textit{(i)} a state preparation routine, \textit{(ii)} the evaluation in parallel of the quantum classifiers and \textit{(iii)} the access to the combined decision. This approach is based on Bayesian Model Averaging (BMA) that exploits many models whose parameters are fixed so as to span a large part of parameters domain.
The strength of this approach is that the individual classifiers do not have to be trained.
However, the algorithm assumes two oracles whose form is not precisely defined in terms of quantum gates. Furthermore, the BMA approach is not very used in ML because of limited performance in real-world applications \cite{domingos1997does,domingos2000bayesian}. In fact, it has been shown that models combination works better by enriching the space of hypotheses, not by approximating a Bayesian model average. On the other side, classical ensemble methods (e.g. Random Forest) generate a collection of complementary hypotheses whose predictions are compatible with the data. These hypotheses are induced by fitting the same model under different training conditions.

Another QML algorithm based on ensemble methods is the idea of \textit{Quantum Boosting} \cite{arunachalam2020quantum}.
In this case, the authors suppose a weak learner $\mathcal{A}$ and try to improve its performance by simulating adaptive boosting procedure \cite{freund1999short} that allows converting a weak learning algorithm to a strong one. This is done by achieving a quadratic improvement over classical AdaBoost.
The main limitation in this case is the ability to prepare efficiently multiple copies of the same quantum states that encode the training set. Also, it assumes to execute the Quantum Phase Estimation algorithm that is a full-coherent protocol requiring a fault-tolerant quantum computer to be executed.

Finally, the idea of a quantum ensemble based on bagging strategy has been recently investigated \cite{macaluso2020quantum}. In this case, the authors propose a quantum approach to aggregate multiple and diverse functions to obtain the equivalent of a prediction based on bagging in classical ensemble methods using a quantum circuit. However, specific quantum routines to test in practice the effectiveness of this approach are not provided.

\subsection{Contribution}\label{sec: contribution}

The purpose of this work is to provide a general framework to tackle classification problems through quantum ensembles.
In particular, we extend and complete the formulation for quantum ensemble based on bagging strategy \cite{macaluso2020quantum}, describe in detail the implementation of a quantum ensemble, and discuss the possibility of employing the same algorithm for randomisation and boosting.

The high-level idea is to design a quantum algorithm that propagates an input state to multiple quantum trajectories in superposition in such a way that a sum of individual results from each trajectory is obtained.
From a technical point of view, the algorithm is able to generate different transformations of the training set in superposition, each entangled with a quantum state of a control register. Thus, a quantum classifier $F$ is applied to obtain a large number of classifications in superposition. By averaging those predictions, the ensemble prediction can be accessed by measuring a single register.

As a consequence of this convenient architecture, our method implies three main computational advantages.
First, the ensemble size (i.e., number of simple base models) scales exponentially compared to classical methods while increasing the depth of the correspondent quantum circuit linearly, since the proposed quantum ensemble requires only $d$ steps to generate $2^d$ different transformations of the same training set in superposition.
Second, having entangled states entails an additive impact of the single weak classifiers, as opposed to the usual multiplicative burden of classical implementations.
This means that the time cost of implementing the ensemble is not dominated by the cost of the single classifier but rather by the data encoding strategy.
Third, the number of state preparation routines is equivalent to implementing just the single classifier since the classification routine is assumed to work via interference, and its use is propagated to all the quantum trajectories in superposition with just one execution.
In addition, the algorithm also allows obtaining the ensemble prediction by measuring a single register, and it makes the evaluation of large ensembles feasible with relatively small circuits.

Finally, we conduct experiments on simulated and real-world data by defining a simple classification routine based on cosine distance to be used as single weak learner.

\section{Quantum Algorithm for Classification Ensemble} \label{sec: quantum ensemble}


In this section we introduce the basic idea of our quantum algorithm for ensemble classification using bagging  in the context of binary classification. The boosting and randomisation approaches, instead, are  discussed in Section \ref{sec:other_ensemble}.

The algorithm adopts 
three quantum registers: data, control, test. 
The \textit{data} register encodes the training set and it is employed together with the $d$-qubits \textit{control} register
to generate $2^d$ altered copies of the training set in superposition. 
The \textit{test} register, instead, encodes unseen observations from the test set.
%
Starting from these three registers, the algorithm involves four main steps: \textit{state preparation}, \textit{sampling in superposition}, \textit{learning via interference} and \textit{measurement}.

\paragraph{(Step 1) State Preparation \\}\label{paragraph: State Preparation}
\noindent
\textit{State preparation} consists in the initialisation of the \textit{control} register into a uniform superposition through a Walsh-Hadamard gate and the encoding of the training set $(x,y)$ in the $data$ register: 
%
%
\begin{align}\label{eq: state preparation}
    \ket{\Phi_0} = \big(W \otimes S_{(x,y)}\big)\overset{d}{\underset{i=1}{\otimes}} \ket{0} \otimes \ket{0} = 
    \big( H^{\otimes d} \otimes S_{(x,y)} \big) \overset{d}{\underset{i=1}{\otimes}} \ket{0} \otimes \ket{0} = \overset{d}{\underset{i=1}{\otimes}} \ket{c_i} \otimes \ket{x,y},
\end{align}
where $S_{(x,y)}$ is the state preparation routine for the training set and it strictly depends on the encoding strategy, $W$ is the Walsh-Hadamard gate and $\ket{c_i}$ is the $i$-th qubit of the control register initialised into a uniform superposition between $\ket{0}$ and $\ket{1}$
%



\paragraph{(Step 2) Sampling in Superposition \\}\label{paragraph: Sampling in Superposition}
The second step regards the generation of $2^d$ different transformations of the training set in superposition, each entangled with a state of the $control$ register. To this end, $d$ steps are necessary, where each step consists in the entanglement of the $i$-th control qubit with two transformations of $\ket{x,y}$ based on two random unitaries,  $U_{(i,1)}$ and $U_{(i,2)}$, for $i=1, \dots, d$.
The most straightforward way to accomplish this is to apply the $U_{(i,j)}$ gate through controlled operations, using as control state the two basis states of the current control qubit. In particular, the generic $i$-th step 
involves the following three transformations:

\begin{itemize}
        \item 
        First, the controlled-unitary $CU_{(i,1)}$ is executed to entangle the transformation $U_{(i,1)}\ket{x,y}$ with the excited state of the $i$-th control qubit: 
            \begin{align}
        \ket{\Phi_{i,1}}= & \Big(CU_{(i,1)}\Big) \ket{c_i} \otimes \ket{x,y} \nonumber \\ = & \Big(CU_{(i,1)}\Big) \frac{1}{\sqrt{2}}\big(\ket{0} + \ket{1} \big) \otimes \ket{x,y} \nonumber \\
                        = & \frac{1}{\sqrt{2}}\Big(\ket{0}\ket{x,y} + \ket{1} U_{(i,1)}\ket{x,y} \Big)
        \end{align}
        %
        \item
        Second, the $i$--th control qubit is transformed based on Pauli--$X$ gate:
            \begin{align}
                \ket{\Phi_{i,2}} = & ( X \otimes \mathbb{1}) \ket{\Phi_{i,1}}
                                \nonumber \\ = & \frac{1}{\sqrt{2}}\Big(\ket{1}\ket{x,y} + \ket{0} U_{(i,1)}\ket{x,y} \Big)
                \end{align}
        \item
        Third, a second controlled-unitary $CU_{(i,2)}$ is executed:
                \begin{align}
                \ket{\Phi_{i}}= & \Big( CU_{(i,2)}\Big)\ket{\Phi_{i,2}}  \nonumber \\ = & \Big( CU_{(i,2)}\Big) \frac{1}{\sqrt{2}}\Big(\ket{1}\ket{x,y} + \ket{0} U_{(i,1)}\ket{x,y} \Big) \nonumber \\
                = & \frac{1}{\sqrt{2}}\Big(\ket{1}U_{(i,2)}\ket{x,y} + \ket{0} U_{(i,1)}\ket{x,y} \Big).
    \end{align}

\end{itemize}
These three transformations are repeated for each qubit in the control register and, at each iteration, two random $U_{(i,1)}$ and $U_{(i,2)}$ are applied. 
After $d$ steps, the \textit{control} and \textit{data} registers are fully entangled and $2^d$ different quantum trajectories in superposition are generated (more details are provided in the Appendix \ref{appendix: Quantum Ensemble as Simple Averaging}).
The output of this procedure can be expressed as follows:
%
%
\begin{align}
        \ket{\Phi_{d}}
        = & \frac{1}{\sqrt{2^d}} \sum_{b = 1}^{2^{d}} \ket{b} V_b\ket{x,y} = \frac{1}{\sqrt{2^d}} \sum_{b = 1}^{2^{d}} \ket{b}\ket{x_b,y_b}
\end{align}
%
where $V_b$ results from the product of $d$  matrices $U_{(i,j)}$ and it represents a single quantum trajectory which differ from the others for at least one matrix $U_{(i,j)}$.
In general, it is possible to refer to the unitary $V_b$ as a unitary that transforms the original training set to obtain a random sub-sample of it:
\begin{equation}\label{eq:single_subsample}
  \ket{x,y} \xrightarrow{V_{b}} \ket{x_b, y_b}. 
\end{equation}
The composition of $V_b$ strictly depends on the encoding strategy choosen for data. In Section \ref{sec:quantum_cosine_classifier} we provide an example of $U_{(i,j)}$ based on the qubit encoding strategy, where a single observation is encoded into a qubit. Notice that, the only requirement to perform ensemble learning using bagging effectively is that small changes in the product of the unitaries $U_{(i,j)}$ imply significant differences in $(x_b, y_b)$,  since the more independent sub-samples are, the better the ensemble works.
%




\paragraph{(Step 3) Learning via Interference \\}\label{paragraph: Learning via Interference}
The third step of the algorithm is \textit{Learning via Interference}. First, the \textit{test} register is initialised to encode the test set, $x^{(\text{test})}$, considering also an additional register to store the final predictions:
\begin{align}
    (S_{x^{(\text{test})}} \otimes \mathbb{1}) \ket{0}\ket{0} =\ket*{x^{(\text{test})}}\ket{0}.
\end{align}
%
Then, the $data$ and $test$ registers interact via interference to compute the estimates of the target variable. To this end, 
we define a quantum classifier $F$ that satisfies the necessary conditions described in Section \ref{sec:ensemble_learning}.
In particular, $F$ acts on three registers to predict $y^{(\text{test})}$ starting from the training set $(x_b, y_b)$:
\begin{equation}\label{eq:oracle_classifier}
     \ket{x_b, y_b}\ket*{x^{(\text{test})}}\ket{0}  \xrightarrow{F} \ket{x_b, y_b} \ket*{x^{(\text{test})}}\ket*{\hat{f}_b}.
\end{equation}
Thus, 
$F$ represents the classification function $\hat{f}$ that estimates the value of the target variable of interest. 
For example, in binary classification problems, 
the prediction can be encoded into the probability amplitudes of a qubit, where the state $\ket{0}$  encodes one class, and the state $\ket{1}$  the other. 

The \textit{Learning via Interference} step leads to:
\begin{align}\label{eq:classification via interference}
    \ket{\Phi_{f}} 
                & = \Big(\mathbb{1}^{\otimes d} \otimes F \Big) \ket{\Phi_d} \nonumber \\ 
                & = (\mathbb{1}^{\otimes d} \otimes F )\Bigg[\frac{1}{\sqrt{2^d}}\sum_{b=1}^{2^d} \ket{b} \ket{x_b, y_b}\Bigg] \otimes
                \ket*{x^{(\text{test})}}
                \ket{0}  \nonumber \\ 
                & =  \frac{1}{\sqrt{2^d}}\sum_{b=1}^{2^d} \ket{b} \ket{x_b, y_b} \ket*{x^{(\text{test})}}\ket*{\hat{f}_b} 
\end{align}
where $\hat{f_b}$ represents the prediction for $x^{(\text{test})}$ given the  $b$-th training set, and it is implemented via quantum gate $F$. 
%
Notice that expressing the prediction according to Equation \eqref{eq:classification via interference} implies that it is necessary to execute $F$ only once in order to propagate its use to all the quantum trajectories. 
Furthermore, as consequence of Steps 2 and 3, the \textit{b-th} state of the $control$ register is entangled with the \textit{b-th} value of $\hat{f}$.

\begin{figure}[ht!]
\scalebox{.9}{\input{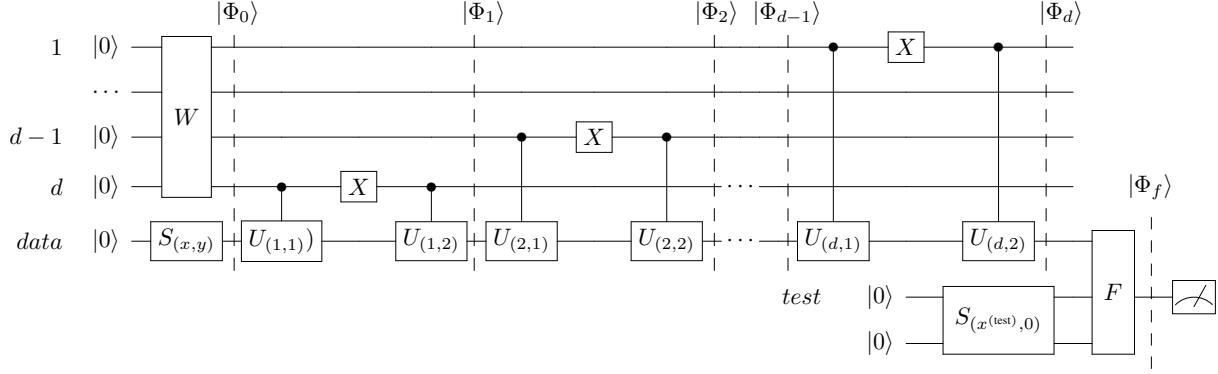}}
\caption{
Quantum algorithm for ensemble classification. The circuit contains $d$ pairs of unitaries $U_{(i,1)}$, $U_{(i,2)}$ and $d$ control qubits. It produces an ensemble of $B$ classifiers, where $B = 2^{d}$. The single evaluation of $F$ allows propagating the classification function $\hat{f}$ in all trajectories in superposition.
The firsts $d$ steps allows generating $B$ transformations of the training set $(x,y)$ in superposition, and each transformation is entangled with a quantum state of the $control$ register (firsts $d$ qubits). Thus,  the test set $x^{(\text{test})}$ is encoded in the $test$ register that interferes with all samples in superposition. 
Finally, the ensemble prediction is obtained as the average of individual results from each trajectory.
}
\label{circuit:quantum_ensemble}
\end{figure}

\paragraph{(Step 4) Measurement \\}\label{paragraph: Measurement}
Measuring the $test$ register allows retrieving the average of the predictions provided by all the classifiers: 
\begin{align}\label{eq: prediction of quantum ensemble}
    \left\langle M \right\rangle & =  \braket{\Phi_f|\mathbb{1}^{\otimes d} \otimes \mathbb{1} \otimes \mathbb{1} \otimes M}{\Phi_f} \nonumber \\
    & =  
    \frac{1}{2^d}\sum_{b=1}^{2^d}\braket{b}{b} \otimes  \braket{(x_b,y_b)}{(x_b,y_b)} \otimes\braket*{x^{(\text{test})}}{x^{(\text{test})}} \otimes\braket*{\hat{f}_b|M}{\hat{f}_b}  \nonumber \\
    & =  \frac{1}{2^d}\sum_{b=1}^{2^d} \braket*{\hat{f}_b|M}{\hat{f}_b} =
    \frac{1}{2^d}\sum_{b=1}^{2^d}\left\langle M_b \right\rangle  \nonumber \\
    & =  \frac{1}{B} \sum_{b=1}^B \hat{f}_b = \hat{f}_{bag}(x^{(\text{test})}|x,y)
\end{align}
where $B = 2^d$ and $M$ is a measurement operator (e.g. Pauli-$Z$ gate).
The expectation value $\left\langle M \right\rangle$ computes the ensemble prediction since it results from the average of the predictions of all the weak learners. Thus, if the two classes of the target variable are encoded in the two basis states of a qubit, it is possible to access to the ensemble prediction by single-qubit measurement:
\begin{align}
    \hat{f}_{bag} = \sqrt{a_0}\ket{0}+\sqrt{a_1}\ket{1} 
\end{align}
where $a_0$  and $a_1$ are the average of the probabilities for $x^{(\text{test})}$ to be classified in class $0$ and $1$, respectively. 
%
The quantum circuit of the quantum ensemble is illustrated in Figure \ref{circuit:quantum_ensemble}.

\subsection{Quantum Algorithm for Boosting and Randomisation}\label{sec:other_ensemble}


The same framework presented above can be adapted with slight variations to allow also randomisation and boosting.
The main principle of the ensemble based on randomisation consists in the introduction of casual perturbations that decorrelate the predictions of individual classifiers as much as possible. In this case, it is possible to loosen the constraints imposed on the classifier $F$, which can be generalised beyond weak learners. The procedure described in Step 2 (\textit{Sampling in Superposition}), indeed, can be employed to introduce a random component in the single learner, so to decrease the accuracy of each individual model. As a consequence, the predictions are less correlated and the variance of the final prediction is reduced.

Technically, it is necessary to define a classification routine  which can be decomposed in the product of $V_b$ and $F$. Here, the different trajectories do not simulate the bootstrap procedure as for bagging, but they are part of the classification routine and introduce randomisation in the computation of $\hat{f}$. 
In practice, we define a unitary $G_b$ that performs the following transformation:
\begin{align} \label{eq:quantum randomisation single gate}
    \ket{x,y}\ket*{x^{(\text{test})}}\ket{0} \xrightarrow{G_b}
    \ket{x,y}\ket*{x^{(\text{test})}}\ket*{\hat{f}_b},
\end{align}
where $G_b=V_b F$ is the quantum classifier composed by $F$ -- common to all the classifiers -- and $V_b$ which is its random component -- different for each quantum trajectory. This formulation allows rewriting the quantum state in Equation \eqref{eq:classification via interference} as:
\begin{align} \label{eq:quantum randomisation}
    \ket{\Phi_{f}} 
    =\frac{1}{\sqrt{2^d}}\sum_{b=1}^{2^d}\ket{b}  G_b  \ket{x,y}\ket*{x^{(\text{test})}}\ket{0} = \frac{1}{\sqrt{2^d}}\sum_{b=1}^{2^d} \ket{b} \ket{x,y}\ket*{x^{(\text{test})}}\ket*{\hat{f}_b}.
\end{align}
%
%

Likewise, the proposed framework can also be adapted for boosting, where the estimates provided by the single classifiers are weighted so that individual models do not contribute equally to the final prediction. In practice, the only difference is that the amplitudes of the control register now need to be updated as the computation evolves. As a result, the output of a quantum ensemble based on boosting can be described as:
%
%
\begin{align} \label{eq:quantum_boosting}
    \ket{\Phi_{f}} 
    = \frac{1}{\sqrt{2^d}}\sum_{b=1}^{2^d} \alpha_b \ket{b} \ket*{\hat{f}_b},
\end{align}
where the contribution of $\hat{f_b}$ to the ensemble depends on $\alpha_b$.
However, although in principle this approach fits in the scheme of a boosting ensemble,
the difficulty in updating the $control$ register is non-trivial.  

To summarise, the main difference between quantum bagging and the other approaches is the way we define the unitaries $U_{(i,j)}$ and $F$. However, the exponential growth  that comes from the advantage of generating an ensemble of $B=2^d$ classifiers in only $d$ steps still holds.
\subsection{Aggregation Strategy and Theoretical Performance}
\begin{figure}[ht]
\centering
\includegraphics[scale =.5]{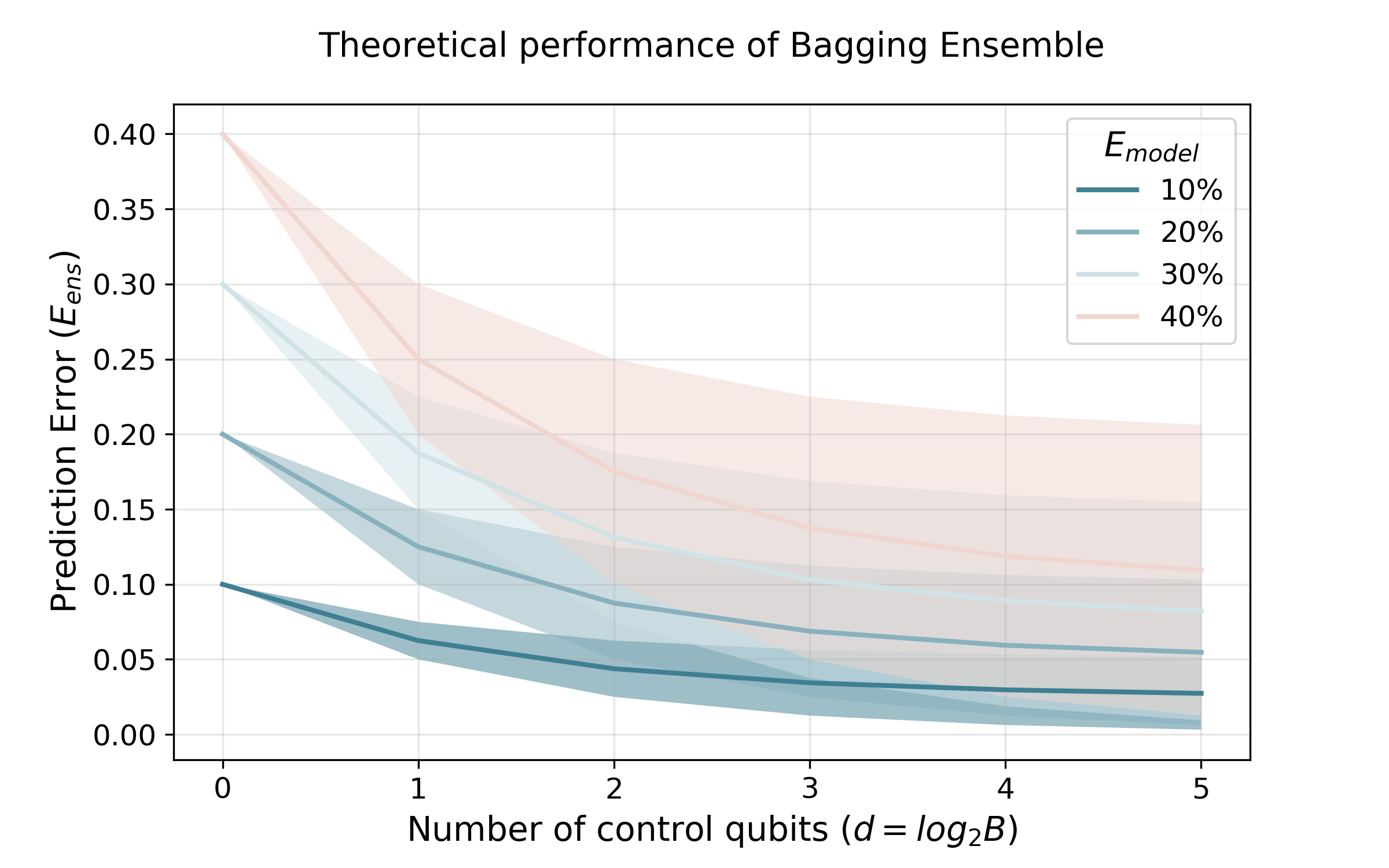} 
\caption{Theoretical performance of the quantum ensemble based on the expected prediction error of the base classifiers ($E_{\text{model}}$) and their average correlation $(\rho)$. 
The ensemble size depends on the number of qubits $d$ in the control register. Each solid line corresponds to an error level, with coloured bands obtained by varying $\rho$  between $0$ (lower edge) and $0.5$ (upper edge).}
\label{fig:theory_bagging_performance}
\end{figure}

When considering classical implementations of ensemble algorithms, it is possible to distinguish two broad families of methods based on the strategy adopted to aggregate the predictions of the individual models.
On one hand, the most popular technique used in ensemble classification is \textit{majority voting}, 
where each classifier votes for a target class and the most frequent is then selected.
On the other hand, an alternative strategy is given by \textit{simple averaging}. In this case, the target probability distribution provided by individual models is considered, and the final prediction is computed  as follows: 
%
\begin{align}
    f_{\text{avg}}^{(i)}(x) = \frac{1}{B}\sum_{b=1}^B f_b^{(i)}(x),
\end{align}
where $B$ is the ensemble size and $f_b^{(i)}(x)$ is the probability for $x$ to be classified in the $i$-th class provided by the $b$-th classifier. This approach allows a reduction of the estimates variance  \cite{tumer1996analysis} and has shown good performance even for large and complex datasets \cite{xu1992methods}.
%
In particular,
the error $E_{\text{ens}}$ of an ensemble obtained averaging $B$ individual learners can be expressed as 
\cite{jacobs1995methods, oza2008classifier}:
\begin{align}\label{eq:error ensemble}
    E_{\text{ens}} = \frac{1+\rho(B-1)}{B}E_{\text{model}}
\end{align}
where $E_{\text{model}}$ is the expected error of the single models and  $\rho$ is the average correlation among their errors. 
Hence, the more independent the single classifiers are, the greater the error reduction due to averaging.
A graphical illustration of the theoretical performance of an ensemble as a function $B$, $\rho$ and $E_{\text{model}}$ is reported in Figure \ref{fig:theory_bagging_performance}.

%
%
%
Coming to our implementation of the quantum ensemble, the prediction of the single classifier is encoded into the probability amplitudes of a quantum state and the final prediction is computed by averaging the results of all quantum trajectories in superposition. Implicitly, this means that the quantum ensemble fits in the simple averaging strategy. 
Thus, the possibility to generate exponentially larger ensembles at the cost of increasing linearly the number of control qubits $d$ allows quantum ensemble to improve significantly the performance of the single classifier (Figure \ref{fig:theory_bagging_performance}) using relatively small circuit ($d \sim 10$).


\subsection{Computational Complexity}\label{sec:discussion}

Classically,
given a number $B$ of base learners  and a dataset $(x_i,y_i)$ for $i=1, \dots N$, where $x_i$ is a $p$-dimensional vector and $y_i$ is the target variable of interest,  the overall time complexity for training an ensemble based on randomisation or bagging scales at least linearly with respect to $B$ and polynomially in $p$ and $N$: 
\begin{align*}
    \underbrace{\mathcal{O}(BN^{\alpha}p^{\beta})}_\text{Training} + \underbrace{\mathcal{O}(Bp)}_\text{Testing} \qquad \alpha, \beta \geq 1,
\end{align*}
where $\alpha$ and $\beta$ depend on the single base model and $N^{\alpha}p^{\beta}$ is its training cost. 
In boosting, instead, the model evolves over time and the individual classifiers are not independent. This usually implies higher time complexity and less parallelism.

Despite this clear definition of the computational cost, comparing the classical algorithm to its quantum counterpart is not straightforward since they belong to different classes of complexity.
For this reason, we benchmarked the two approaches by looking at how they scale in terms of the parameters of the ensemble, i.e, the ensemble size $B$ and the cost of each base model.
In particular, this resolves in considering the Boolean circuit model \cite{arora2009computational} for the classical ensemble, and the depth of the corresponding quantum circuit for the quantum algorithm.
In light of this definition, the quantum algorithm described in Section \ref{sec: quantum ensemble} is able to generate an ensemble of size $B=2^d$ in only $d$ steps.
This means that, assuming a unitary cost for each step, we are able to increase exponentially the size of the ensembel while increasing linearly the depth of the correspondent quantum circuit. 
Furthermore, the cost of the single classifier  is additive -- instead of multiplicative as in classical ensembles -- since
it is necessary to execute the quantum classifier $F$ only once to propagate its application to all quantum trajectories in superposition, as shown in Equation \eqref{eq:classification via interference}. In addition, the cost of the state preparation routine is equivalent to any other quantum algorithm for processing the same training and test sets.
However, this comparison does not take into account  the additional cost due to  state preparation which is not present in classical ensembles. Also, the quantum ensemble comes with an extra cost related to the implementation of the gates  $U_{(i,j)}$, that strictly depends on the encoding strategy chosen for the data and needs to be evaluated for a any specific implementation.
\section{Experiments}\label{se}

To  test how our framework for quantum ensemble works in practice,  we  implemented  the  circuit  illustrated in Figure \ref{circuit:quantum_ensemble} using IBM qiskit \cite{Qiskit}.
Then, we conducted experiments on simulated and real-world dataset to show that $(i)$ one execution of a quantum classifier allows retrieving the ensemble prediction, and that $(ii)$ the ensemble outperforms the single model.

\subsection{Quantum Cosine Classifier}\label{sec:quantum_cosine_classifier}

In order to implement the quantum ensemble, a  classifier that fulfils the conditions in  Equation \eqref{eq:oracle_classifier} is necessary. For this purpose, we define a simple routine for classification based on the swap-test \cite{PhysRevLett.87.167902}  that stores the cosine distance between two vectors into the amplitudes of a quantum state.
This metric describes how similar two vectors are depending on the angle that separates them, irrespectively of their magnitude.
The smaller the angle between two objects, the higher the similarity. 
Starting from this, the high-level idea is predicting a similar target class for similar input features. In particular, for any test observation $(x^{(\text{test})}, y^{(\text{test})})$ we take one training point $(x_{b}, y_{b})$ at random and we express the probability of $y^{(\text{test})}$ and $y_{b}$ being equal as a function of the similarity between $x^{(\text{test})}$ and $x_{b}$:
\begin{align}\label{equation:cosine classifier}
    Pr\left(y^{(\text{test})} = y_{b}\right) = \frac{1}{2}+\frac{\left[d\left(x_{b}, x^{(\text{test})}\right)\right]^2}{2}
\end{align}
where 
$d(\cdot, \cdot)$ is the cosine distance between $x_{b}$ and $x^{(\text{test})}$.
Thus, the final classification rule becomes:
 \begin{align} \label{classification_rule}
    y^{(\text{test})} = 
    \begin{cases} y_{b}, & \mbox{if } Pr\left(y^{(\text{test})} = y_{b}\right) > \frac{1}{2} \\ 1- y_{b}, 
                      & \mbox{otherwise }  
    \end{cases}
\end{align}

Notice that, by definition, $Pr\left(y^{(\text{test})} = y_{b}\right)$ is bounded in $[\frac{1}{2}, 1]$, which means that Equation \eqref{classification_rule} will always estimate the same class as the training point,  unless $x_{b}$ and $x^{(\text{test})}$ are orthogonal. As a consequence, the cosine classifier performs well only if the test and training observations happen to belong to the same target class.

\begin{figure}
    \centering
\includegraphics[scale=.3]{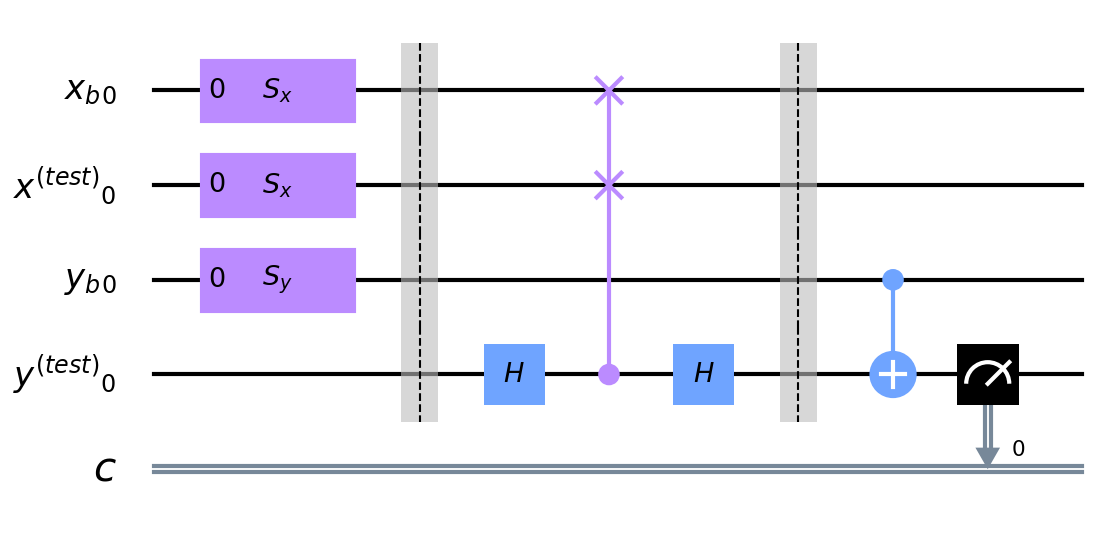}    \caption{Quantum circuit of the cosine classifier using $x_{b}$ as training vector and $x^{(\text{test})}$ as test vector. The training label $y_{b}$ is either $\ket{0}$ or  $\ket{1}$ based on the binary target value. The measurement of the qubit $y^{(\text{test})}$ provides the prediction for the test observation whose features are encoded in $x^{(\text{test})}$.} 
\label{circuit:quantum_cosine}
\end{figure}

The quantum circuit that implements the cosine classifier is reported in Figure \ref{circuit:quantum_cosine}.
It encodes data into three different registers: the training vector $x_{b}$, the training label $y_{b}$ and the test point $x^{(\text{test})}$. An additional qubit is then used to store the prediction. The algorithm is made of three steps. First, data are encoded into three different quantum registers through a routine $S$.  Second, the swap-test transforms the amplitudes of the qubit $y^{(\text{test})}$ as a function of the squared cosine distance. In particular, after the execution of the swap-test the probability of getting the basis state $\ket{0}$ is between $1/2$ and $1$, hence
 the probability of class $0$ is never lower than the probability of class $1$.
Third, a controlled Pauli-$X$ rotation is applied using as control qubit the label of the training vector. This implies that $y^{(\text{test})}$ is left untouched if  $x_{b}$ belongs to the class $0$. Otherwise,
the amplitudes of the $y^{(\text{test})}$ qubit are inverted,
and $Pr(y^{(\text{test})} = 1)$ becomes higher as the similarity between the two vectors increases. 
%

\begin{figure}
    \centering
\includegraphics[scale=.6]{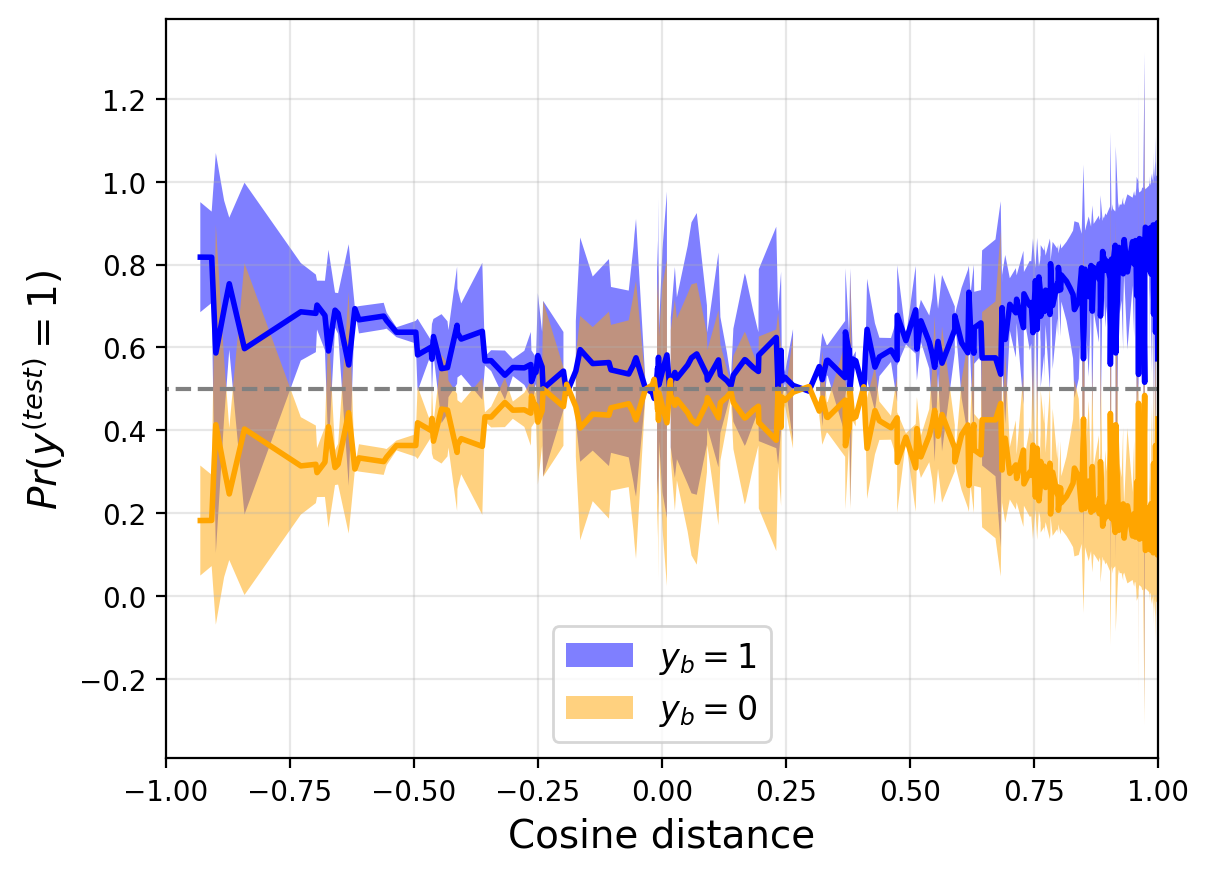}    
\caption{Predictions of the cosine distance classifier based on $10^3$ randomly generated datasets per class. The classifier is implemented using the circuit in Figure \ref{circuit:quantum_cosine} on a $7$-qubit quantum device (\textit{ibmq\_casablanca}). The same implementation assuming a perfect quantum device is reported in Appendix \ref{appendix: Quantum Cosine Classifier}, Fig. \ref{fig:multiple_run_avg_qasm}.}
\label{fig:quantum_cosine_results}
\end{figure}

Thus, the quantum cosine classifier performs classification via interference and it allows calculating the probability of belonging to one of the two classes by single-qubit measurement. A detailed description of the quantum cosine classifier is provided in Appendix \ref{appendix: Quantum Cosine Classifier}.
Furthermore, it is a weak method
with high-variance, since it is sensitive to the random choice of the training observation. 
In addition, it requires data to be encoded using qubit encoding, where a dataset with $N$ $2$-dimensional observations $x_{b}$ is stored into $N$ different qubits. This allows the definition of $U_{(i,j)}$ for the quantum ensemble in terms of random swap gates that move observations from one register to another. 
 All these features make this classifier a good candidate for ensemble methods.

\subsection{Quantum Ensemble as Simple Averaging}\label{Quantum Ensemble as Simple Averaging}
As a proof-of-concept for the quantum ensemble based on
bagging, we consider different $20$ random generated datasets, each containing four training points ($2$-dim features and label) and one test example. For every simulated dataset, each training point is used as training observation and fed into a quantum cosine classifier as input so to provide an estimate for a test observation $x^{\text{(test)}}$.
Thus, the quantum ensemble, which requires only one execution of the quantum cosine classifier is executed. The quantum circuit of the ensemble uses two qubits in the $control$ register ($d=2$) and eight in the $data$ register, four for the training vectors $x_b$ and four for training labels $y_b$. Two additional qubits are then used for the test observation, $x^{(\text{test})}$,  and the final prediction.
Notice that the four matrices $U_{(i,j)}$ need to be fixed to guarantee that each quantum trajectory $V_b$ described in Section \ref{sec: quantum ensemble} provides the prediction of different and independent training points.

For each training point, the quantum cosine classifier is implemented using the quantum cosine classifier, and then a prediction for the test point is calculated.  This small experiment aims to prove that the quantum ensemble prediction is exactly the average of the values of all trajectories in superposition and it can be obtained with just one execution of the classification routine.
 \begin{figure}[!ht]
    \centering
    \includegraphics[scale=.4]{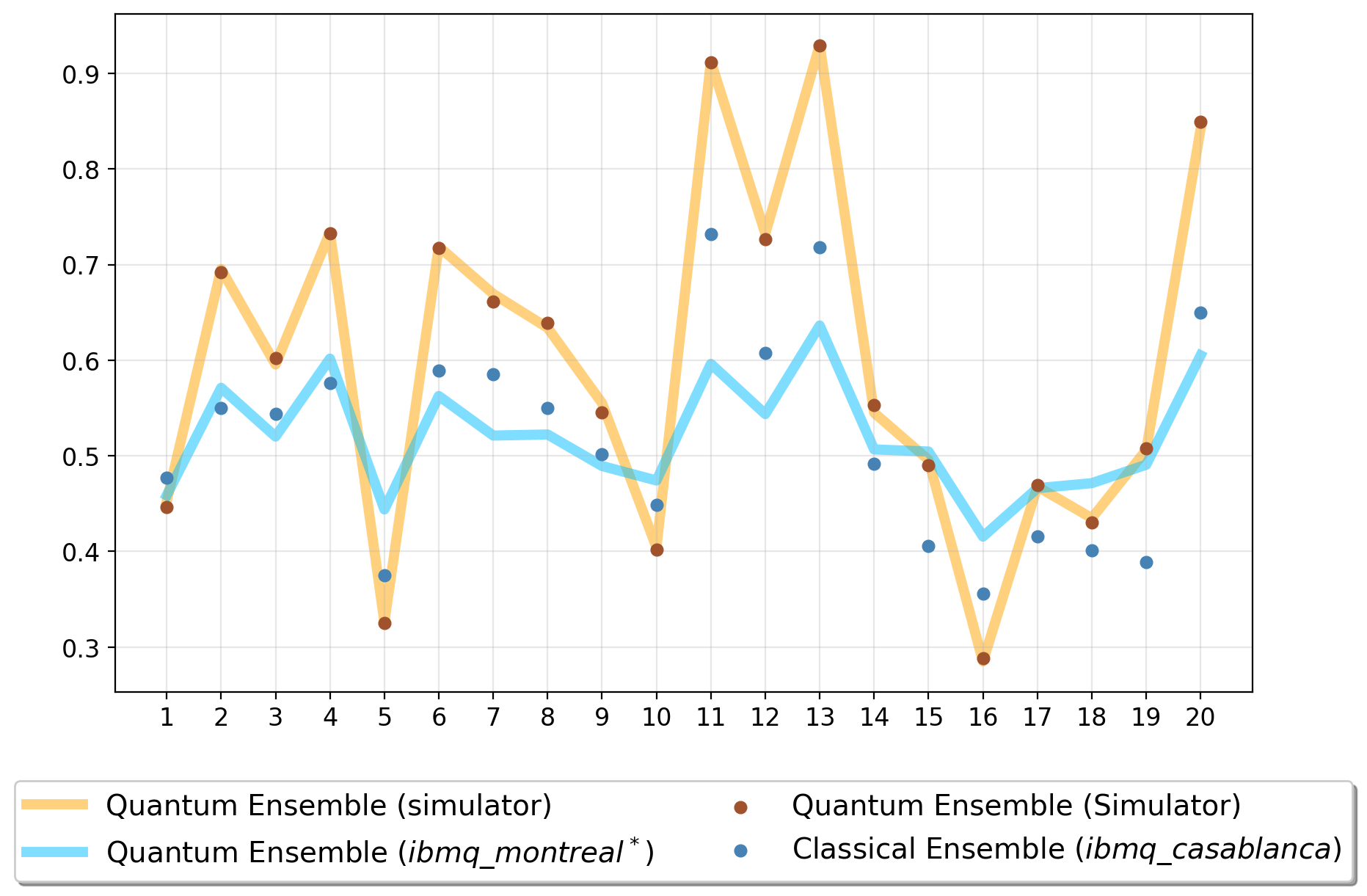}
\caption{Comparison between the Quantum Ensemble and the Classical Ensemble as a result of the classical average of four quantum cosine classifiers executed separately. Both approaches are performed on a simulator (orange line, brown dots) and on real device (light blue line, blue dots). Importantly,  the Quantum Ensemble implementation on real device is performed using noisy simulations of the specific quantum device ($ibmq\_16\_montreal^*$). These  simulations are only an approximation of the real errors that occur on actual devices, but still allow us to test the effectiveness of the quantum ensemble on near-term devices. The details about the implementation  in terms of quantum gates of the quantum ensemble is reported in Appendix \ref{appendix: Quantum Ensemble as Simple Averaging}}
\label{fig:multiple_run_avg}
\end{figure}
 
Results are shown in Figure \ref{fig:multiple_run_avg}. The agreement between the quantum ensemble (orange line) and the average (brown dots) is almost perfect, which confirms the possibility to perform quantum ensemble with the advantages described in Section \ref{sec: quantum ensemble} in a fault-tolerant setting. Results considering the real device (light blue line) show slightly deterioration, this may be due to the depth of the quantum circuit which seems to be prohibitive considering current available quantum technology.


\subsection{Performance of the Quantum Ensemble in a Fault-Tolerant Setting}

\begin{figure}[ht]
    \centering
    \includegraphics[scale=.6]{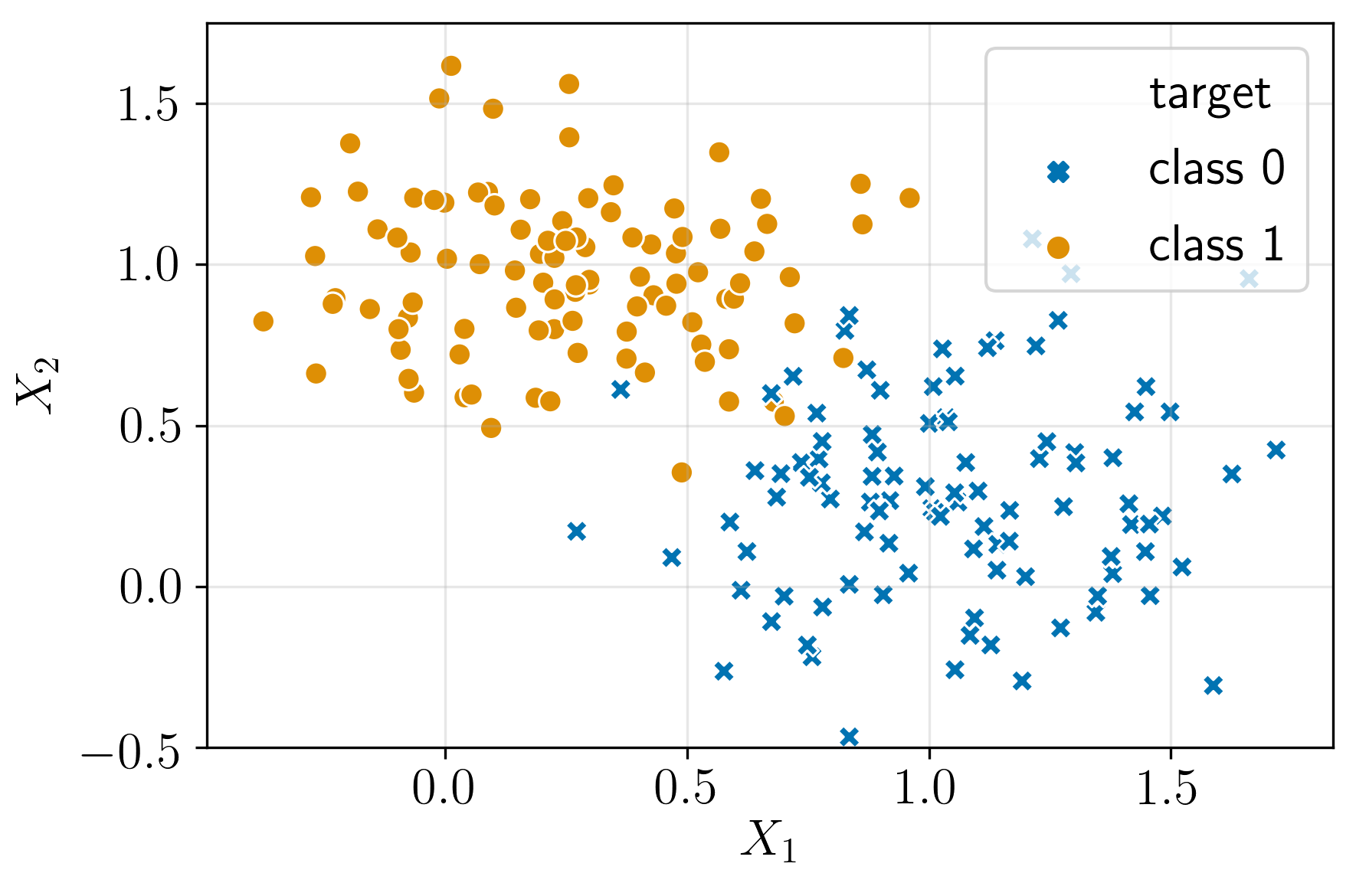}
\caption{Dataset generated by two independent bivariate Gaussian distributions. Mean vectors for the two classes are 
  $(1, 0.3)$ 
  and 
  $(0.3, 1)$.
  The two distributions have the same diagonal covariance matrix, with constant value of $0.3$. }
  \label{fig:dataset}
\end{figure}

To show that the quantum ensemble outperforms the single classifier we generated a simulated dataset and compared the performance of the two models (the pseudo code of the ensemble is described in the Appendix \ref{algorithm: quantum ensemble}).
In particular, we drew a random sample of $200$ observations ($100$ per class) from two independent bivariate Gaussian distributions, with different mean vectors and the same covariance matrix (Figure \ref{fig:dataset}). Then, we used the $90\%$ of the data for training and the remaining $10\%$ for testing.

Notice that the definition of the cosine classifier, implies executing the classification routine once for each test point (See Appendix \ref{appendix: Quantum Ensemble as Simple Averaging} for more details).
We considered two performance metrics, accuracy and Brier Score (BS). The accuracy is the fraction of labels predicted correctly by the quantum model, and it is evaluated using a set of observations not employed for training (test set). Instead, the BS measures the difference between the probability estimates and the true label in terms of mean squared error:
\begin{align}
    BS = \frac{1}{N_{\text{test}}} \sum_{i=1}^{N_{\text{test}}} \left[y_i - f(x_i) \right]^2 ,
\end{align}
where $N_{\text{test}}$ is the number of observation in the test set, $y_i$ and $f(x_i)$ are, respectively, the true label and the probability estimates provided by the quantum model for the $i^{th}$ observation (for quantum ensemble see Equation \eqref{eq: prediction of quantum ensemble}). Hence a low BS score implies a good prediction. 
Due to the randomness introduced by choice of training points, we repeated the experiments $10$ times and evaluated the classifiers in terms of the mean and standard deviation of both accuracy and BS. The experiments of this section are all run assuming a perfect quantum device. 
Results are shown in Table \ref{table:results}.

\begin{table}[ht!]
\begin{tabular}{ccc|cc|cc} \toprule
& & & \multicolumn{2}{c}{\textbf{Accuracy}} & \multicolumn{2}{c}{\textbf{Brier Score}} \\ \cmidrule(lr){4-7}
    {\textbf{d}} & {\textbf{B}} & {\textbf{N}} & {\textbf{Mean}} & {\textbf{Std dev}} & {\textbf{Mean}} & {\textbf{Std dev}} \\ \midrule
    \rowcolor{Gray}  0 &  1  & 1 & .55 & .09 & .21 & .05 \\
                     1 &  2  & 2  & .92 & .09 & .14 & .09 \\
    \rowcolor{Gray}  2 &  4  & 4 & .91 & .09 & .15 & .05 \\
                     3 &  8  & 8  & .96 & .04 & .14 & .04\\
    \rowcolor{Gray}  4 &  16  & 8  & .98 & .02 & .13 & .02  \\ \bottomrule
\end{tabular}
   \caption{Performance comparison between quantum cosine classifier and quantum ensemble of different sizes $B = 2^d$. The first row indicates the performance of the single quantum cosine classifier. The column \textbf{N} indicates the number of training points used to build the ensemble, that is limited to $8$ because of limited number of qubits that is possible to simulate.}%
  \label{table:results}
\end{table}
The single quantum cosine classifier performed only slightly better than random guessing, with an average accuracy of $55 \%$. Yet, the quantum ensemble managed to achieve definitely better results, with both metrics improving as  the ensemble size grows.

\begin{figure}[!ht]
    \begin{subfigure}{0.45\textwidth}
        \includegraphics[width=65mm]{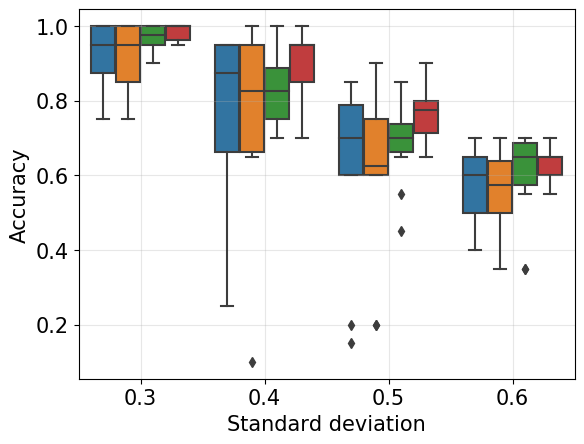}
        \label{fig:overlapp_acc}
    \end{subfigure}
    \begin{subfigure}{0.45\textwidth}
    \includegraphics[width=65mm]{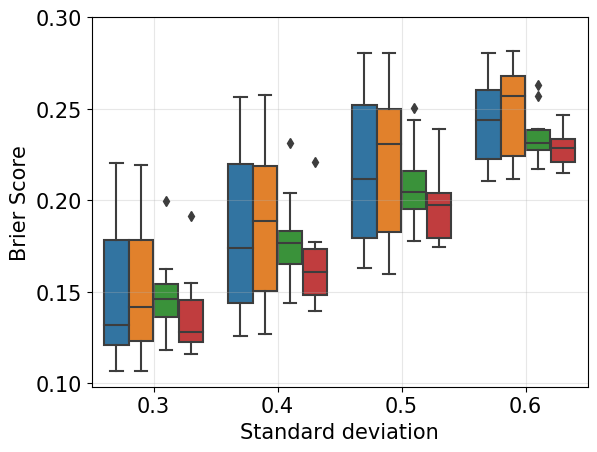} 
     \label{fig:overlapp_brier}
    \end{subfigure}
    \centering
        \includegraphics[scale=.7]{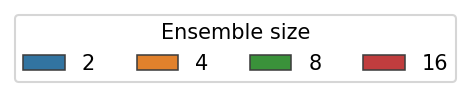}
    \caption{Distribution of the performance metrics as a function of the ensemble size (legend colors) and the separation between the two classes ($x$ axis).}
    \label{fig:overlapp_all}
 \end{figure}

\paragraph{Variance Analysis \\}
In addition, we investigated how the quantum ensemble behaves as the generated distributions get closer and less separated. To this end, we drew multiple samples from the two distributions, each time increasing the common standard deviation so to force reciprocal contamination. Results are reported in Figure \ref{fig:overlapp_all}. 
The accuracy showed a decreasing trend as the overlap of the distributions increased. The opposite behaviour  is observed for the Brier Score.
Also, the shape of the boxplots is much narrower for greater ensemble sizes (green and red boxplots) than for smaller ones (blue and orange). Hence, this confirms that the variability of the ensemble decreases as the number of weak learners adopted grows, as expected.

\subsection{Benchmark on real-world datasets}

In this section we test the performance of the quantum ensemble on real-world datasets that are usually employed to benchmark classical machine learning algorithms.

\subsubsection{Datasets description}\label{cap5: dataset description}

The simulation of a quantum system on a classical device is a challenging task, even for systems of moderate size. For this reason, experiments consider only datasets with a relatively small number of observations ($100$--$150$) that will be split in training ($90\%$) and test ($10\%$) set.  Furthermore, in order to limit the overall number of qubits, the  Principal Components Analysis (PCA) is performed to reduce the number of features to $2$. 
For all the datasets, a given subset of training points (\textbf{N}) are encoded in the $data$ register and then the prediction retrieved by measuring the label qubit of the test register. This is performed for each test point and finally the test error in terms of Accuracy and BS is considered to evaluate the generalisation error of the quantum model. The following four different datasets will be considered. 

\paragraph{MNIST} It is a large dataset of handwritten digits that is commonly used to benchmark various image processing systems. In particular, it is usually employed for training and testing different algorithms in the field of machine learning. 
The original dataset contains $60.000$ black and white images, each represented by $28 \times 28$ pixels. Thus, the single image can be described as a vector of binary $784$ features ($0$ if the pixel is white, $1$ if it is black).  Also, each image belongs to a class of ten possible, that represents the digit depicted in the image. The current implementation of the cosine classifier is arranged to solve a binary classification problem. Hence, only two different classes will be considered, the digits $0$ and $9$.
\paragraph{Iris}
The Iris flower data set collects the data to quantify the morphologic variation of Iris flowers of three related species \cite{anderson1936species}. 
The data set consists of 50 examples from each of three species of Iris (Iris $setosa$, Iris $virginica$ and Iris $versicolor$). Four features describe each observation: the length and the width of the sepals and petals, in centimetres. 
The dataset is often used in statistical learning theory as classification and clustering examples to test algorithms.
Since the current implementation of the quantum ensemble solve a binary classification problem, only two species at dataset will be considered (three different datasets in total).

\subsubsection{Results}

The results of the quantum ensemble on the real-world datasets are reported in Table \ref{fig:ensemble_benchmark_all}. For each dataset, the quantum ensemble is implemented simulating a perfect quantum device. 

\begin{table}
\centering
\begin{tabular}{cc|cc|cc|cc|cc} \toprule
& &  \multicolumn{2}{c}{\textbf{Iris (0 vs 1)}} & \multicolumn{2}{c}{\textbf{Iris (0 vs 2)}} & \multicolumn{2}{c}{\textbf{Iris (1 vs 2)}} & \multicolumn{2}{c}{\textbf{MNIST}} \\ \cmidrule(lr){3-10}
    {\textbf{d}} & {\textbf{B}/\textbf{N}} & {\textbf{Accuracy}} & {\textbf{BS}} & {\textbf{Accuracy}} & {\textbf{BS}} & {\textbf{Accuracy}} & {\textbf{BS}} & {\textbf{Accuracy}} & {\textbf{BS}} \\ \midrule
    \rowcolor{Gray}  0 &  1  & .49 & .284 & .49 &  .284 &  .49 & .445 & .50 & .337 \\
                     1 &  2  & 1.0 & .137 & 1.0 & .276 & .51 & .240 & .79 & .209 \\
    \rowcolor{Gray}  2 &  4  & 1.0 & .138 & 1.0 & .139 & .52 & .240 & .78 & .208 \\
                     3 &  8  & 1.0 & .136 & 1.0 & .138 & .61 & .241 & .84 & .197 \\ \bottomrule
\end{tabular}
    \caption{Performance of the quantum ensemble on real-world datasets.}
    \label{fig:ensemble_benchmark_all}
\end{table}

Comparing the results in terms of the ensemble size $(B=2^d)$, it is possible to observe a decreasing trend of the BS and an increasing trend for accuracy. This confirms the ability of the quantum ensemble to improve the performance of the single quantum classifier. However, the quantum ensemble do not achieve good performance in the case of the dataset \textbf{Iris (1 vs 2)}. 

\section{Conclusion and Outlook}

In this paper, we propose a quantum framework for binary classification using ensemble learning. The correspondent algorithm allows generating a large number of trajectories in superposition, performing just one state preparation routine. Each trajectory is entangled with a quantum state of the control register and represents a single classifier.
This convenient design allows scaling exponentially the number of base models with respect to the available qubits in the control register ($B=2^d$). As a consequence, we can obtain an exponentially large number of classification while increasing only linearly the depth of the correspondent quantum circuit with respect to the size of the control register.
Furthermore, when considering the overall time complexity of the algorithm, the cost of the weak classifier is additive, instead of multiplicative as it usually happens. 

In addition, we present a practical implementation of the quantum ensemble using bagging where the quantum cosine classifier is adopted as base model. In particular, we show experimentally that the ensemble prediction corresponds to the  average of all the probabilities estimated by the single classifiers. Moreover, we test our algorithm on synthetic and real-world (reduced) datasets and demonstrate that the quantum ensemble systematically outperforms the single classifier. Also, the variability of the predictions decreases as the we add more base models to the ensemble.

However, the current proposed implementation requires the execution of the classifier for just one test point at the time, which is a big limitation for real-world applications. 
In this respect, the main challenge to tackle in order to make the framework effective in the near future is the design of a quantum classifier  based on interference that guarantees a more efficient data encoding strategy (e.g. amplitude encoding) and that is able to process larger datasets. Nevertheless, these upgrades would imply a different definition of $U_{(i,j)}$ for the generation of multiple and diverse training sets in superposition. Further, the design of a more accurate base quantum model is  necessary. 

Another natural follow-up is the implementation of quantum algorithms for randomisation and  boosting. In this work, we only referred to an ensemble based on bagging  because the learning step was performed independently in each quantum trajectory and the weak classifiers were assumed to be sensitive to perturbations of the training set. However, with appropriate amendments and loosening these constraints, we believe that it is possible to design other types of ensemble techniques.

Finally, it is important to notice that the idea of model aggregation has already been adopted in the context of variational quantum algorithms to build the quantum Single Layer Perceptron \cite{macaluso2020variational}. Thus, it is natural to consider the provided quantum architecture for the quantum ensemble as a natural extension of it if adopted adequately in hybrid quantum-classical computation.

Although some challenges still remain, we believe this work is a good practical example of how Machine Learning, in particular ensemble classification, could benefit from Quantum Computing.

\section*{Acknowledgments}


We acknowledge the use of IBM Quantum services for this work. The views expressed are those of the authors, and do not reflect the official policy or position of IBM or the IBM Quantum team.
We acknowledge the access to advanced services provided by the IBM Quantum Researchers Program.


\bibliographystyle{unsrt}  
\bibliography{references.bib}  

\newpage

\appendix

\appendixhead

\section{Quantum Ensemble as Simple Averaging}\label{appendix: Quantum Ensemble as Simple Averaging}

Here we describe the quantum circuit to obtain four \textit{independent} quantum trajectories in superposition considering a quantum ensemble of cosine classifiers (Section \ref{Quantum Ensemble as Simple Averaging}).

\paragraph{\textbf{(Step 1) State Preparation} \\}

For a 2-qubit $control$ register $(d=2)$, we can build an ensemble of $B=4$ classifiers. The $data$ encodes a single observation using a single qubit. In particular, given a dataset made up of $N$ observations $\{x_i, y_i\}_{i=1, \dots, N}$, where $x_i =(x_{i,1}, x_{i,2})$ is a $2$-dimensional vector and $y_i \in \{0,1\}$ is the binary target variable, the $data$ register encodes $N$ training points $2 \times N$ qubits:
\begin{align}
    \text{data register: } \underset{features}{\left( \overset{4}{\underset{i=1}{\otimes}} \left| x_i \right\rangle \right)}\otimes  \underset{ labels}{\left( \overset{4}{\underset{i=1}{\otimes}} \left| y_i \right\rangle \right)} = \ket{features} \ket{labels},
\end{align}
where the values $x_{i,1}$ and $x_{i,2}$ are encoded into the amplitudes of a single qubit: 
\begin{align}
    \left| x_i \right\rangle = x_{i,1}\left| 0 \right\rangle +  x_{i,2}\left| 1 \right\rangle,
\end{align}
and the two classes of the target variable are represented by the two basis states of a single qubit. Thus, if $\left| y_i \right\rangle=\left| 0 \right\rangle$ the $i$-th observation belongs to the class $0$. Otherwise, if $\left| y_i \right\rangle=\left| 1 \right\rangle$ the $i$-th observation belongs to the class $1$. 

\textit{Qubit encoding strategy} allows to store a training set of $4$ observations using an $8$-qubit $data$ register. 
In formulas, the state preparation step leads to:
\begin{align*}
    \left|\Phi_0\right\rangle &=
    \big( H^{\otimes 2} \otimes S_{(x,y)} \big)\left|0\right\rangle \otimes \left|0\right\rangle \otimes \left|0\right\rangle \nonumber \\ 
        & =
   \left|c_1\right\rangle \otimes \left|c_2\right\rangle \otimes \left|x\right\rangle \left|y\right\rangle \nonumber\\
    & =
    \frac{1}{\sqrt{2}}\big(\left|0\right\rangle+\left|1\right\rangle\big) \otimes \frac{1}{\sqrt{2}}\big(\left|0\right\rangle+\left|1\right\rangle\big) \otimes \left|x_0,x_1,x_2,x_3\right\rangle \left|y_0,y_1,y_2,y_3\right\rangle ,
\end{align*}

where $S_x$ is the routine which encodes in the amplitudes of a qubit a real vector $x$ and $H$ is the Hadamard transformation.

\paragraph{\textbf{(Step 2) Sampling in Superposition}\\}

The second step regards the generation of $2^d$ different transformations of the training set in superposition, each entangled with a state of the control register. To this end, $d$ steps are necessary, where each step consists in the entanglement of the $i$-th control qubit with two transformations of $\left|x,y\right\rangle$ based on two random unitaries, $U_{(i,1)}$ and $U_{(i,2)}$, for $i = 1,2$.

The sampling in superposition step leads to the following quantum state:
\begin{align*}
\left|\Phi_{1}\right\rangle
=  \frac{1}{2}\Big[
 \hspace{.2em} & \left|00\right\rangle U_{(2,1)}U_{(1,1)}\left|x_0,x_1,x_2,x_3 \right\rangle \left| y_0,y_1,y_2,y_3\right\rangle
\nonumber \\ + & 
\left|01\right\rangle U_{(2,1)}U_{(1,2)}\left|x_0,x_1,x_2,x_3\right\rangle \left|y_0,y_1,y_2,y_3\right\rangle
\nonumber \\ + & 
\left|10\right\rangle U_{(2,2)}U_{(1,1)}\left|x_0,x_1,x_2,x_3\right\rangle \left|y_0,y_1,y_2,y_3\right\rangle
\nonumber \\ + & 
\left|11\right\rangle U_{(2,2)}U_{(1,2)}\left|x_0,x_1,x_2,x_3\right\rangle \left| y_0,y_1,y_2,y_3\right\rangle 
         \Big]  .
\end{align*}

In order to obtain independent quantum trajectories, we provide the following definition for $U_{(i,j)}$: 
\begin{align}
    U_{(1,1)} & = \text{SWAP}(x_0,x_2) \times \text{SWAP}(y_0,y_2);  \label{eq:U11} \\
    U_{(1,2)} & = \text{SWAP}(x_1,x_3) \times \text{SWAP}(y_1,y_3); \label{eq:U12}  \\
    U_{(2,1)} & = \mathbb{1};  \label{eq:U21} \\
    U_{(2,2)} &= \text{SWAP}(x_2,x_3) \times \text{SWAP}(y_2,y_3); \label{eq:U22} 
\end{align}
where $ \mathbb{1}$ is the identity matrix. Thus, we get:
\begin{align*}
    \left|\Phi_{2}\right\rangle = \frac{1}{2}\Big[ 
    & \left|11\right\rangle \left|x_0, x_3, x_1, x_2\right\rangle  \left|y_0, y_3, y_1, y_2\right\rangle  
    \\ + &
    \left|10\right\rangle \left|x_2, x_1, x_3, x_0\right\rangle  \left|y_2, y_1, y_3, y_0\right\rangle \nonumber\\
    \hspace{.1em} 
    + &
    \left|01\right\rangle \left|x_0, x_3, x_2, x_1\right\rangle \left|y_0, y_3, y_2, y_1\right\rangle \\
    + &
    \left|00\right\rangle \left|x_2, x_1, x_0, x_3\right\rangle \left|y_2, y_1, y_0, y_3\right\rangle
    \Big].
\end{align*}
We can see that swap operations allows to entangle different observationsstored in the \textit{data} register to different state of the $control$ register. In particular, if considering the last qubit of the \textit{features} and \textit{labels} (sub-)registers, the above choices for $U_{(i,j)}$ guarantee that each quantum state of the control register is entangled with a different training observation, if considerinf the last qubit of the \textit{data} register. Using a compact representation:

\begin{align}
     \left|\Phi_{2^{'}}\right\rangle & = \frac{1}{2}\Big[
    \left|11\right\rangle \ket{\dots} \left|x_2\right\rangle  \left|y_2\right\rangle  
    + 
    \left|10\right\rangle \ket{\dots} \left|x_0\right\rangle\left|y_0\right\rangle 
    +
    \left|01\right\rangle \ket{\dots} \left|x_1\right\rangle\left|y_1\right\rangle 
    +
    \left|00\right\rangle \ket{\dots} \left|x_3\right\rangle \left|y_3\right\rangle 
    \Big] \nonumber \\ & =
    \frac{1}{\sqrt{4}}\sum_{i=0}^{3}\left|i\right\rangle \ket{\dots}\left|x_i,y_i\right\rangle .
\end{align}

Notice that, in this case, the $i$-th basis state does not correspond to the integer representation of the binary state. 

Importantly, when considering random swap operations as unitaries $U_{(i,j)}$ instead of the fixed ones ( Eq. \eqref{eq:U11}, \eqref{eq:U12}, \eqref{eq:U21}, \eqref{eq:U22}) , we have no guarantees that  the four quantum trajectories will be independent, but this randomness is necessary to generate the typical scenario of ensemble methods where the different training sets are randomly sampled from the original one.

\paragraph{\textbf{(Step 3) Learning via interference} \\}

The $test$ register is initialised to encode the test set, $\tilde{x}$, considering also an additional qubit to store the final prediction:
\begin{align}
    (S_{\tilde{x}} \otimes \mathbb{1}) \ket{0} \ket{0} =\ket*{x^{(test)}} \otimes \ket{0} =  \ket*{x^{(test)}} = \left(x_{\text{test},1}\left| 0 \right\rangle +  x_{\text{test},2}\left| 1 \right\rangle \right) \otimes \ket{0}.
\end{align}

Then, the $data$ and $test$ registers interact via interference using the quantum version of the cosine classifier (gate $F$) to compute the estimates of the target variable:

\begin{align*}
    \left|\Phi_{f}\right\rangle 
                = & \Big(\mathbb{1}^{\otimes 2} \otimes F \Big) \left|\Phi_{d}\right\rangle \nonumber \\ 
                = & \frac{1}{\sqrt{4}}\sum_{b=1}^{4} \left|b\right\rangle \left|x_b, y_b\right\rangle\ket*{x^{(test)}} \ket*{\hat{f}_b} .
\end{align*}

Since the $4$ points of the training set are in superposition, the application of the quantum cosine classifier allows computing $4$ different predictions for the test point, $\{\hat{f}_b \}_{b=1, \dots 4}$, executing the classifier only once. 

\begin{figure}[ht]
    \centering
    \includegraphics[width=150mm, height=90mm]{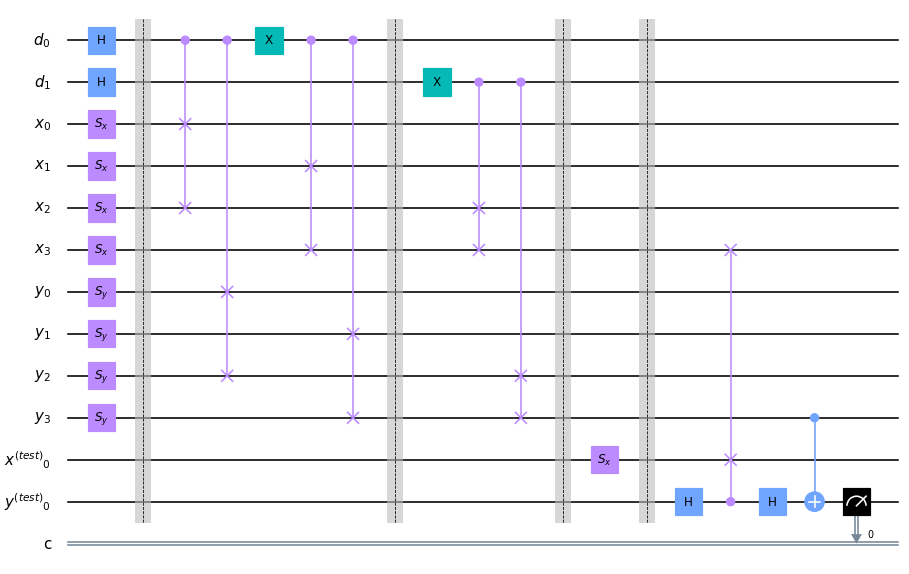}
    \caption{Qiskit implementation of the quantum ensemble for four independent quantum trajectories.}
    \label{fig: quantum circuit for ensemble}
\end{figure}

\paragraph{\textbf{(Step 4) Measurement}\\}
Due to the entanglement between the predictions for $\tilde{x}$ and the $control$ register the expectation measurement allows retrieving the average of all the predictions, which correspond to the ensemble prediction that uses bagging strategy aggregation:
\begin{align*}
    \left\langle M \right\rangle = &  
    \frac{1}{4}\sum_{b=1}^{4} \left\langle\hat{f}_b|M|\hat{f}_b\right\rangle = \frac{1}{4} \sum_{b=1}^4 \hat{f}_b = \hat{f}_{bag}(\tilde{x}|x,y) .
\end{align*}
The implementations of the quantum ensemble to perform simple averaging is depicted in Figure \ref{fig: quantum circuit for ensemble}.

\newpage



\section{Quantum Cosine Classifier}\label{appendix: Quantum Cosine Classifier}

Classically, cosine classifier is defined as follows:
\begin{align}
    Pr\Big(y^{(test)} = y_{b}\Big) = \frac{1}{2}+\frac{\Big[d\big(x_{b}, x^{(test)}\big)\Big]^2}{2}
\end{align}
where $(x_{b}, y_{b})$ is a random training example, $x^{(test)}$ the test point and $d(\cdot, \cdot)$ the cosine distance between $x_{b}$ and $x^{(test)}$.
Since the probability of belonging to a class depends on the squared cosine distance between the two vectors, the maximum dissimilarity occurs when training and test observations are orthogonal. In this case, the cosine classifier  assigns a uniform probability distribution in the two classes for $y^{(test)}$. This means that the cosine classifier performs well only if the test point belongs to the same class of the training point. 

The quantum circuit that implements the cosine classifier (Figure \ref{circuit:quantum_cosine}) encodes data into three different registers: the training vector $x^{(i)}$, the training label $y^{(i)}$ and the test point $x^{(test)}$. One last qubit is used to store the prediction. 

The algorithm is made of the following three steps.

\paragraph{Step 1: State Preparation\\} 

The state preparation routine can be performed independently for each qubit:

\begin{align}
    \left|\Phi_1\right\rangle = \Big( S_{x_{b}} \otimes S_{x^{(test)}} \otimes S_{y_{b}} \otimes  \mathbb{1}  \Big) \left|0\right\rangle \left|0\right\rangle \left|0\right\rangle \left|0\right\rangle  = \left|x_{b}\right\rangle \ket*{x^{(test)}} \left|y_{b}\right\rangle \left|0\right\rangle ,
\end{align}

where $S_{x}$ is the routine which encodes in the amplitudes of a qubit a $2$-dimensional, normalised real vector $x$.

\paragraph{Step 2: Execution of the swap test\\}

In the second step, the swap-test transforms the amplitudes of the qubit $y^{(test)}$ as a function of the squared cosine distance:

\begin{align}
    \left|\Phi_2\right\rangle =  &
    \left(\mathbb{1} \otimes \mathbb{1} \otimes \mathbb{1} \otimes H\right)
    \left(\text{cswap} \otimes \mathbb{1} \otimes C \right) 
    \left(\mathbb{1} \otimes \mathbb{1} \otimes \mathbb{1} \otimes H\right) 
    \ket{x_{b}} \ket*{x^{(test)}} \ket{y_{b}} \ket{0} \nonumber \\  
   = & 
   \frac{1}{2}
   \left[
   \left( \ket{x_{b}}\ket*{x^{(test)}} + \ket*{x^{(test)}}\ket{x_{b}} \right)\ket{y_{b}}\ket{0} + 
   \left(\ket{x_{b}}\ket*{x^{(test)}} - \ket*{x^{(test)}}\ket{x_{b}} \right)\left|y_{b}\right\rangle  \left|1\right\rangle \right) ,
\end{align}

where $H$ is the Hadamard gate, CSWAP is the controlled-swap operation which uses the last qubit (position of gate $C$) as control qubit to swap $\ket{x_{b}}$ and $\ket{x^{(test)}}$.
After the execution of the swap test the probability to readout the basis state $\ket{0}$, that is the probability for the test observation to be classified in class $0$ is:
\begin{align}
    P(y^{(test)} = \left|0\right\rangle) = 
    \frac{1}{2}+\frac{|\langle x_{b}|x^{(test)} \rangle|^2}{2} .
\end{align}

\paragraph{Step 3: Controlled Pauli-$X$ gate\\}
The third step consists of applying a controlled-Pauli-$X$ gate using as control qubit the label of the training vector. This implies that $y^{(test)}$ is left untouched if $x_{b}$ belongs to the class $0$. Otherwise, the amplitudes of the $y^{(test)}$ qubit are exchanged, and the probability $P(y^{(test)} = 1)$ is higher as the similarity between the two vectors increases.
\begin{align}
    \ket{\Phi_3} = \left(\mathbb{1} \otimes \mathbb{1} \otimes \text{C-X} \right) \ket{\Phi_2} .
\end{align}
At this point the expectation measurement provides on the last qubit provides the prediction of interest.

\begin{figure}[!ht]
    \centering
    \includegraphics[scale=.6]{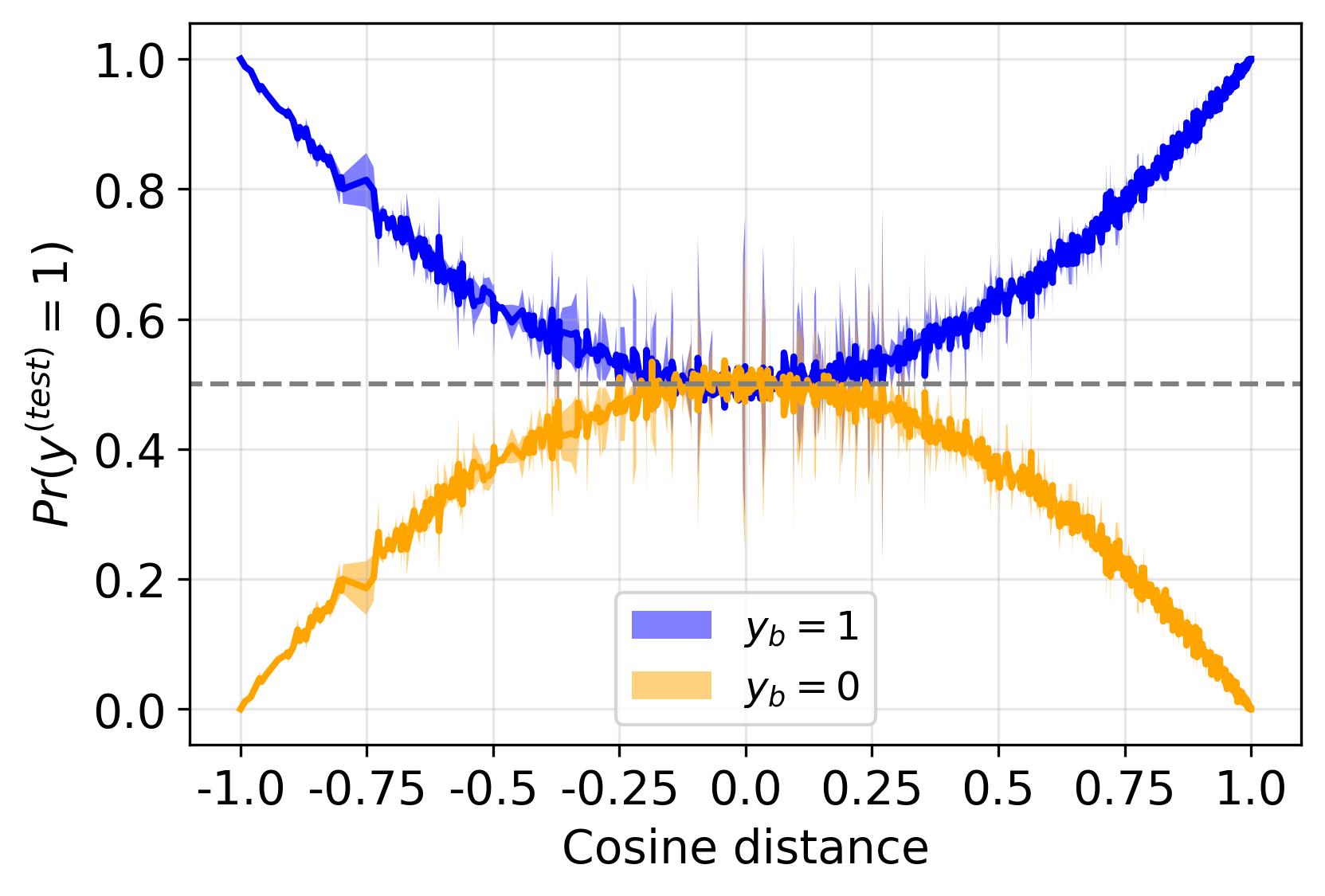}
\caption{Predictions of the cosine distance classifier based on $10^3$ randomly generated datasets per class. The classifier is implemented using the circuit in Figure \ref{circuit:quantum_cosine}.}
\label{fig:multiple_run_avg_qasm}
\end{figure}

The result predictions of the quantum cosine classifier assuming a perfect quantum device are depicted in Figure \ref{fig:multiple_run_avg_qasm}.

\newpage

\section{Algorithm for Quantum Ensemble}\label{algorithm: quantum ensemble}
 This section presents the implementation of the quantum ensemble to produce the results shown in Section \ref{sec: quantum ensemble}.

\begin{algorithm}[H]
            \SetAlgoLined
            \KwResult{Predictions of the binary target value for all points in the test set}
             \vspace{0.5em}
             \textbf{Input:} 
             
              - $2n$--qubit data register, $d$--qubit control register, $2$-qubit test register 
              
              - Pauli-Z (measurement) operator $\langle \sigma_z \rangle$
              
              \vspace{1em}
            
            \For{each point $\tilde{x}$ in the test set}{
            
             \vspace{1em}
 
             \# \textit{(Step 1) State Preparation}
             
             \vspace{.5em}
             
             Encode $n$ random training points into the $n \times 2$ qubits of the $data$ register: 
             $(x_1, y_1),  \dots, (x_n, y_n)  \xrightarrow{S_{(x,y)}} \ket{x_1, \dots, x_n; \hspace{.3em} y_1, \dots, y_n } = \ket{features; \hspace{.3em} labels}$

             Initialise the $d$ qubits of $control$ register into a uniform superposition: $\ket{0\dots 0} \xrightarrow{W} \frac{1}{\sqrt{2^d}}\sum_{k=0}^{2^d-1} \ket{k}$

             Initialise the $test$ register: $\ket{0, 0} \xrightarrow{S_{(\tilde{x},0)}} \ket{\tilde{x},0}$
             
              \vspace{1.5em}
              
            \# \textit{(Step 2) Sampling in superposition}
            
             \vspace{0.5em}
            
             \For{each qubit in the control register $(i=1, \dots d)$}{
              
              \vspace{0.5em}
              
              Select two pairs of random integers $l,m$ and $l^{'}, m^{'}$ between $1$ and $n$\;
              
              \vspace{1em}
              
              C-SWAP$\left(control(i), features(l), features(m)\right)$\;
              
              C-SWAP$\left(control(i), labels(l), labels(m)\right)$\;
              
              \vspace{0.5em}
              
              Apply Pauli-X gate to the current $control$ qubit \;
              
              \vspace{0.5em}
              
              C-SWAP$\left(control(i), features\left(l^{'}\right), features\left(m^{'}\right)\right)$\;
              
              C-SWAP$\left(control(i), labels\left(l^{'}\right), labels\left(m^{'}\right)\right)$\;
              
             }
             \vspace{1.5em}
             
             \# \textit{(Step 3) Learning via Interference}
             
             \vspace{0.5em}
             
            
            Apply the quantum cosine classifier (gate $F$) using as training set a random pair of qubits (\textit{features, labels}) of the $data$ register\;   
            
          \vspace{1.5em}

            \# \textit{(Step 4) Measurement}
            
            \vspace{0.5em}

            Measure the $test$ register using $\langle \sigma_z \rangle$ operator
            
            }
            
            \textbf{Output:} Ensemble predictions for all points in the test set\;
             \caption{Quantum ensemble of quantum cosine classifiers}
\end{algorithm}

\newpage

\end{document}